# TALCS: AN OPEN-SOURCE MANDARIN-ENGLISH CODE-SWITCHING CORPUS AND A SPEECH RECOGNITION BASELINE

*Chengfei Li, Shuhao Deng, Yaoping Wang, Guangjing Wang, Yaguang Gong, Changbin Chen, Jinfeng Bai*

TAL Education Group, Beijing, China
{lichengfei, dengshuhao1, wangyaoping, wangguangjing, gongyaguang, chenchangbin, baijinfeng1}@tal.com

## Abstract

This paper introduces a new corpus of Mandarin-English code-switching speech recognition—TALCS corpus, suitable for training and evaluating code-switching speech recognition systems. TALCS corpus is derived from real online one-to-one English teaching scenes in TAL education group, which contains roughly 587 hours of speech sampled at 16 kHz. To our best knowledge, TALCS corpus is the largest well labeled Mandarin-English code-switching open source automatic speech recognition (ASR) dataset in the world. In this paper, we will introduce the recording procedure in detail, including audio capturing devices and corpus environments. And the TALCS corpus is freely available for download under the permissive license[1]. Using TALCS corpus, we conduct ASR experiments in two popular speech recognition toolkits to make a baseline system, including ESPnet and Wenet. The Mixture Error Rate (MER) performance in the two speech recognition toolkits is compared in TALCS corpus. The experimental results implies that the quality of audio recordings and transcriptions are promising and the baseline system is workable.

## 1. Introduction

In recent years, due to the development of machine learning and neural network, monolingual speech recognition, such as Mandarin speech recognition and English speech recognition, has achieved great success. Code-switching (CS) is the phenomenon where speakers alternate between different languages within an utterance or between consecutive utterances. There is increasing research interest in developing code-switching automatic speech recognition (CS-ASR) [1] systems as most of the off-the-shelf systems are monolingual and cannot handle code-switched speech. Intuitively, we can build a CS-ASR system simply by means of merging two languages. However, such a CS-ASR system can rarely produce optimal recognition results. This is because there are no within-an-utterance code-switching samples, and the resulting ASR system frequently fails to recognize those code-switching utterances.

One of the main reasons for the success of monolingual speech recognition is that a large number of labeled data are used to train ASR models. However, large industrial datasets are often inaccessible, especially for academic community. Open Speech and Language Resources (Open SLR) project is established to alleviate this problem. For Mandarin ASR, industrial-sized datasets, such as AISHELL-1 [2], provides 100+ hours of open source data and a speech recognition baseline system. WENETSPEECH [3] also provides a multi-domain mandarin corpus for speech recognition, which consists of 10000+ hours high-quality labeled speech, 2400+ hours weakly labeled speech, and about 10000 hours unlabeled speech, with 22400+ hours in total. These open source datasets effectively promote the development of Mandarin speech recognition. For English ASR, LibriSpeech corpus [4] provides about 960 hours of open source data and a speech recognition baseline system. Based on Librispeech corpus, a large number of scholars have proposed various methods to reduce the word error rate (WER) of speech recognition, which has greatly promoted the progress of English speech recognition. And the People's Speech corpus [5] provides a large-scale diverse English speech recognition dataset for academic and commercial usage, which contains 30000 hours of free download data. Both Mandarin ASR and English ASR benefit from these large-scale open source data and have made great progress in academia and industry.

However, it is not an easy thing for CS due to the shortage of corresponding CS corpus in both academic community and industry. To our best knowledge, there are several CS datasets which is freely available. SEAME corpus [6] provides about 112 hours Mandarin-English code-switching conversational speech data, which is collected from residents of Malaysia and Singapore. English-Mandarin code-switching speech recognition contest sponsored by DataTang [7] in China released two training datasets, namely, 200 hours of English-Mandarin code-switching data, and 500 hours of Mandarin data of which there are about 15 hours of similar code-switching data. And this means that the largest open-source CS corpus is only about 215 hours. Actually, for the code-switching dataset, as far as we know, most companies are also lack of data because they do not have relevant business scenarios. In addition, some CS data cannot be open due to unauthorized use, and legal and ethical issues surrounding open source data have also led to a shortage of CS data. For these reasons, CS-ASR is identified as one of the most challenging tasks for ASR systems, either for hybrid ASR systems or end-to-end ASR systems. In order to alleviate the shortage of CS open source data and promote the progress of CS-ASR technology, we present TALCS corpus in this paper. It is released by TAL education group, containing more than 100 speakers and over 580 hours of Mandarin-English code-switching speech data. TALCS is one of the most representative scenes of CS—teacher English teaching scene, and it is also the known largest open-source Mandarin-English code-switching corpus in the world. This corpus is freely available under the permissive license. The TALCS corpus is divided into three parts: training set, development set and test set. Using TALCS corpus, we conduct automatic speech recognition experiments in two popular speech recognition toolkits to make a baseline system, including ESPnet [8] and Wenet [9]. And the Mixture Error Rate (MER) performance of the two speech recognition toolkits is compared in TALCS corpus. The experimental results implies that the quality of audio recordings and transcriptions are promising and the baseline system is workable.

---

[1] https://ai.100tal.com/dataset

This paper is organized as below. Section 2 presents the CS-ASR related work. Section 3 describes the structure of the TALCS corpus, and presents the recording procedure. In Section 4, we describe the process we used to build the CS-ASR system. Finally, in Section 5, we summarizes our work and looks forward to the future work.

## 2. Related work

In ASR community, the sequence-to-sequence acoustic modeling has attracted great attention. In the past ten years, deep neural network (DNN) [10] has gradually replaced Gaussian mixture model (GMM) for acoustic modeling, and the hybrid ASR system composed of acoustic model, language model and lexicon model has achieved convincing performance. Recently, end-to-end (E2E) ASR models such as Connectionist Temporal Classification (CTC) [11], the recurrent neural network transducer (RNN-T) [12], transformer [13] or conformer transducer [14], attention-based encoder-decoder models [15] have gained popularity and achieved state-of-the-art performance in accuracy and latency. In contrast to conventional hybrid ASR systems, they joint learn acoustic and language modeling in a single neural network that is E2E trained from labeled data. For example, the recently proposed Transformer-based end-to-end ASR architectures use deeper encoder-decoder architecture with feedforward layers and multi-head attention for sequence modeling, and comes with the advantages of parallel computation and capturing long-contexts without recurrence.

CS occurs commonly in everyday conversations in multilingual societies. For example, Mandarin and English are often mixed in conversations in some countries of Southeast Asia, such as Malaysia and Singapore, while Cantonese and English are mixed in colloquial Cantonese in Hong Kong [2]. In addition, the CS phenomenon occurs more frequently in the workplace. To handle CS speech, there have been many studies in acoustic modeling [16], language modeling [17, 18], and ASR systems [19-21]. And Mandarin-English CS-ASR is typically a low-resource task due to the scarce acoustic and text resources that contain CS. In CS-ASR, to solve the data scarcity problem during acoustic modeling, one solution is to simply merge two different monolingual datasets into one CS dataset for the training of CS-ASR model, another effective solution is to adapt two well-trained high-resource language acoustic model to the target low-resource domain using transfer learning. These solutions can alleviate the problem of data scarcity to some extent, but the performance of CS-ASR model will reach the bottleneck and cannot be further improved because its training data does not fully match the situation in the actual application scenario.

In order to further improve the performance of CS-ASR, large-scale training data that can match the actual application scenario is still crucial. In order to alleviate the scarcity of CS open source data and promote the progress of CS-ASR technology, we release TALCS corpus, a 580+ hours mono-channel Mandarin-English code-switching speech corpus designed for various code-switching speech processing task. And in the next section, we will introduce the recording procedure and subsequent processing of TALCS corpus in detail.

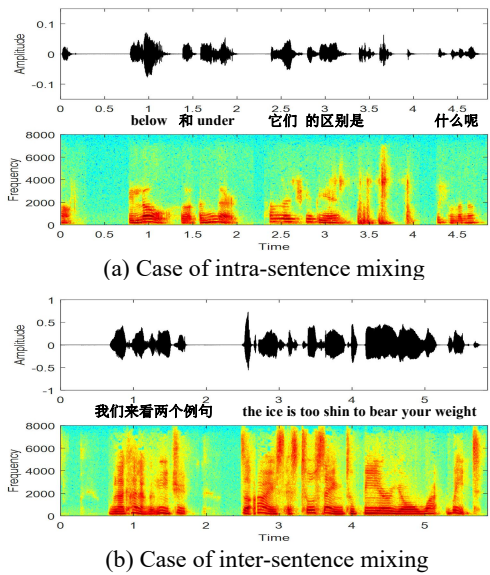

Figure 1: *Cases of two categories of Mandarin-English code-switching in TALCS corpus.*

## 3. Corpus description

TALCS corpus is a subset of the TAL-ASR corpus ( https://ai.100tal.com/dataset), which is a 580+ hours mono-channel Mandarin-English code-switching speech corpus designed for various code-switching speech processing task.

### 3.1. Corpus profile

TALCS corpus comes from online one-to-one teaching scene, in which teachers and students come from different regions of China. And all of the speech utterances of TALCS corpus are recorded by the personal computer microphone. During the course of online one-to-one class, the dialogue between teachers and students is natural, not read aloud, and this means that TALCS corpus is a typical corpus derived from real scenes rather than deliberately produced, which is very conducive to CS-ASR in real scenes. In the online one-to-one course of English teaching, teachers and students mainly use Mandarin for dialogue, and their English pronunciation is relatively standard. TALCS corpus is derived from an online one-to-one course, and teachers and students do not communicate directly face-to-face, so the speech utterances of teachers and students can be recorded through different microphones. For this reason, TALCS corpus has no audio mixing between teachers and students. Considering the privacy of students and corresponding legal and ethical issues, TALCS corpus only includes the speech utterances of teachers.

Generally speaking, Mandarin-English code-switching can be divided into two categories: one is intra-sentence mixing, and another is inter-sentence mixing. As shown in Figure 1, TALCS corpus covers both two cases. Figure 1 (a) shows an example of intra-sentence mixing in TALCS corpus, and it means what is the distinguish between "below" and "under". And we can see that the CS phenomenon occurs three times in this short sentence, which is a typical English-Mandarin-English-Mandarin example in TALCS corpus. Since TALCS corpus comes from online one-to-one English teaching scenes, teachers need to explain the meaning of each word or the distinguish between different words to students, and there are many examples of such intra-sentence mixing. Figure 1 (b) shows an example of inter-sentence mixing in TALCS corpus, and it means let's look at some examples, "the ice is too shin to bear your weight". In online one-to-one English teaching scenes, teachers whose mother tongue is Mandarin often explain some English sentences to students to help them understand some commonly used English expressions, such as tense, active voice, passive voice and so on. Therefore, the phenomenon of inter-sentence mixing often occurs in the online English teaching scene. TALCS corpus contains a large number of intra-sentence mixed audio and inter-sentence mixed audio, and it can well match the CS phenomenon in daily life, which is conducive to the training of CS-ASR model and promote the development of CS-ASR.

The distance between the teacher and the microphone acquisition device is about 20 cm to 30 cm. The TALCS corpus chooses high fidelity microphone audio data and resample it to 16 kHz, 16-bit WAV format, which is the mainstream setup for commercial ASR systems. There are more than 100 teachers in the recording. For the TALCS corpus, the gender is balanced. All teachers are between the ages of 22 and 28 and come from different regions of China. Our English teachers, whose mother tongue is Chinese, have rich experience in English teaching, and their English pronunciation is quite standard. Thus, these teachers can easily switch between English pronunciation and Chinese pronunciation. In addition, the teachers will also be required to pass some grade examinations, such as CET-4 and CET-6, to ensure pronunciation standards. Therefore, the quality of TALCS corpus is reliable. For the entire TALCS corpus, in order to conduct ASR experiments as a baseline system conveniently, we divide it into three parts: training set, development set and test set. The details are presented in Section 3.2 and Section 3.3.

### 3.2. Corpus Transcription

TALCS corpus covers the knowledge points of middle school English, such as grammar, reading comprehension and writing. The corpus including 370K sentences with a length ranging from 0.3 seconds to 30 seconds. Because the audio data comes from one-to-one online course, the environment background noise of corpus is relatively low, and teachers' lecture place are generally in dormitories, families and classrooms. Raw texts are manually filtered to eliminate improper contents involving sensitive political issues, user privacy, pornography, violence, etc. Symbols such as "<", ">", "[", "]", """, "/", "\", "=", etc., are removed. There are no spaces between Chinese characters, and there are spaces between English words. All English words and letters are capitalized and all text files are encoded in UTF-8.

In order to ensure the high quality of TALCS corpus and enable it to be effectively applied to CS-ASR tasks, the label acquisition of TALCS corpus includes two stages: data annotation stage and quality checking stage. In data annotate stage, due to the particularity and complexity of code-switching, there is more demanding for data annotators. In order to ensure the quality of annotating, we have looked for data annotator who have passed CET-4 to annotate the TALCS corpus. And we annotate the audio data of the whole online class and the average duration of each class is about 2 hours. The data annotator divides the 2-hour the audio data into many small sentences according to semantics, and transcribes each small sentence at the same time. Considering the complexity of semantics and different understanding of semantics for different data annotator, only one data annotator is assigned to each class for annotation. It is required to transcribe all the sentences that can be heard clearly, and abandon the sentences that have obvious mispronunciations. And text normalization (TN) is carefully applied towards English words, letter, numbers, name are:

• 1, 2, 3 are normalized to "一", "二", "三".

• All the English words contained are capitalized with spaces between words.

• English letters such as "ING", "ED" are presented in uppercase and also with spaces between letters.

In quality checking stage, TALCS corpus is divided into many parts. Data quality inspectors are asked to check the speech data and transcript of each part at a rate of 20%. Utterances with inconsistent raw text and transcription are marked. If the marked sentences exceed the specified proportion of 10%, the annotating team was asked to reannotate all the data of this part. In addition, data quality inspectors are also asked to check the corpus contents involving student sensitive political issues, such as name, school and so on.

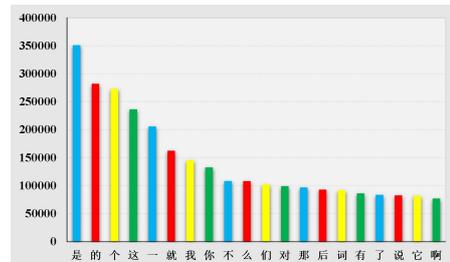

(a) Top 20 Mandarin characters in TALCS corpus

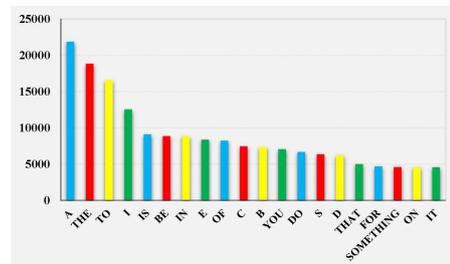

(b) Top 20 English words or letters in TALCS corpus

Figure 2: *The number distribution of 20 Mandarin characters and 20 English words or letters that appear most frequently in TALCS corpus.*

Table 1: *Data structure of TALCS corpus*

| Subset | Number of sentences | Duration (h) |
| --- | --- | --- |
| Training set | 350000 | 555.9 |
| Development set | 5000 | 8.0 |
| Test set | 15000 | 23.6 |
| Total | 370000 | 587.5 |

### 3.3. Data structure

As mentioned above, TALCS corpus comes from the online one-to-one English teaching scene of TAL Education Group. We annotate the audio data of the whole class and the average duration of each class is about 2 hours. In the online one-to-one English teaching course, the teachers need to use their mother tongue, that is Mandarin, to explain the relevant knowledge of English. Therefore, not all of the sentences contain code-switching phenomenon, but the content is closely related. The TALCS corpus contains 3670 commonly used Mandarin characters and 18087 English words and letters. The total number of Mandarin characters and English is about 7746905 and 616936 respectively, with a proportion of about 13/1. Since the Mandarin-English code-switching data are Mandarin-dominated, this proportion is reasonable and consistent with the proportion of Mandarin-English in the phenomenon of CS in daily life. Figure 2 shows the number distribution of 20 Mandarin characters and 20 English words or letters that appear most frequently in TALCS corpus. The TALCS corpus was divide into three parts: training set, development set and test set. Table 1 shows data structure of TALCS corpus, which includes the information of the number of sentences and total duration of each part of the TALCS corpus. And TALCS corpus contains 370000 sentences with a total duration of 587 hours from more than 100 teachers. The training set consists of 350000 sentences with a total duration of about 556 hours, and the development set consists of 5000 sentences with a total duration of about 8 hours, and the test set consists of 15000 sentences with a total duration of about 24 hours.

## 4. Experiments

In this section, using TALCS corpus, we will conduct CS-ASR experiments in two popular speech recognition toolkits, including ESPnet and Wenet. And the experiments have two purposes, one is to prove the reliability of the data set, and another is to provide a feasible baseline system for code-switching speech recognition.

### 4.1. Experimental setup

The Kaldi toolkit [22] is employed to extract 80-dimensional Filter banks and 3-dimensional pitch features, which is used to train the CS-ASR model in both ESPnet toolkit and Wenet toolkit. The window length is set to 30ms and the frame shift is set to 20ms. All the input features are normalized to zero mean and unit variance. We use the ESPnet toolkit and the training set of the TALCS corpus to generate our lexicon and Byte Pair Encoding (BPE) model. In this way, the generated lexicon contains 11296 modeling units, and the out-of-vocabulary (OOV) words are mapped into <unk>.

In the ESPnet toolkit, we conduct the CS-ASR experiments in a conformer model, which uses multi-head attention mechanism to capture the global information of speech signal, and employs convolution module to capture the local information of speech signal, and the ASR model based on conformer structure is proved to have excellent performance. In our CS-ASR experiments, the CS-ASR model consists of a 12-block Conformer encoder and a 6-block Transformer decoder. Both in Conformer encoder and Transformer decoder, the multi-head number and head adim is set to 8 and 512, respectively. The forward linear feedback unit is set to 2048. The CNN kernel is set to 15 in Conformer encoder. The loss function of our CS-ASR model is a logarithmic linear combination of cross entropy loss ($\lambda=0.7$) and CTC loss ($\lambda=0.3$). And the optimizer of our CS-ASR model is Adam. The SpecAugment with 2 frequency masks (F=30) and 3 time masks (T=40) are applied on the input features and the convolution sub-sampling rate of input features is set to 4. The total training epochs is set to 100, and we average 10 best checkpoints in development set as the final model. Following this parameter setting, we train two end-to-end CS-ASR models, a mask CTC CS-ASR model and an autoregressive CS-ASR model for decoding.

In the Wenet toolkit, we conduct the CS-ASR experiments in a U2 model, which unifies streaming and non-streaming end-to-end speech recognition in a single model. The CS-ASR model in the Wenet toolkit has the same model structure and parameter setting as the CS-ASR recognition model in the ESPnet toolkit. And the difference is that in ESPnet, the input feature normalization is used as a data preprocessing step, while in Wenet, the input feature normalization is processed online. The total training epochs is also set to 100, and we average 10 best checkpoints in development set as the final model. We employ four different decoding strategies in the Wenet toolkit for decoding, namely CTC greedy search, CTC prefix beam search, attention and attention rescoring respectively.

Table 2: *The MER performance of different CS-ASR model in TALCS-corpus*

| Model | Dev set | Test set |
|---|---|---|
| Mask CTC (ESPnet) | 9.81 | 9.90 |
| Autoregressive (ESPnet) | **7.30** | **7.34** |
| CTC greedy search (Wenet) | 10.89 | 10.82 |
| CTC prefix beam search (Wenet) | 10.81 | 10.74 |
| Attention (Wenet) | 9.20 | 9.33 |
| Attention rescoring (Wenet) | 9.24 | 9.21 |

### 4.2. Experimental results

The Mixture Error Rate (MER) [23] is used to evaluate the performance of two popular speech recognition toolkits, ESPnet and Wenet, in CS-ASR corpus. And MER takes Mandarin characters and English words as tokens to calculate the edit distance. Table 2 reports the MER performance of the CS-ASR model in ESPnet toolkit and the CS-ASR model in Wenet toolkit in the development set and test set of TALCS-corpus. As the Table 2 shows, the autoregressive CS-ASR model in ESPnet toolkit gets the best MER performance both in the development set and test set of TALCS corpus. And other CS-ASR models also achieved about 10% MER performance both in the development set and test set of TALCS-corpus. Considering that TALCS corpus comes from real online one-to-one teaching scene and CS-ASR is a quite challenging task than monolingual speech recognition, this result can indicate that the TALCS corpus has a high transcription quality. And it should be noted that, since we aim to provide a feasible baseline system for CS-ASR, this result might not reflect the state-of-the-art performance.

## 5. Conclusions

An open-source Mandarin-English code-switching corpus, TALCS corpus, is released in this paper, which comes from real online one-to-one teaching scene. To our best knowledge, it is the largest academically free dataset for Mandarin-English code-switching recognition tasks. Using TALCS corpus, we conduct CS-ASR experiments in two popular speech recognition toolkits, including ESPnet and Wenet, and provide a feasible baseline system for code-switching speech recognition. The experimental results indicate that the TALCS corpus has a high transcription quality and is a challenging dataset for CS-ASR tasks.